\documentclass[runningheads]{llncs}
\usepackage{graphicx}
\usepackage{xcolor}
\usepackage{hyperref}
\usepackage{amsfonts}
\usepackage{array,booktabs}
\newcolumntype{P}[1]{>{\centering\arraybackslash}p{#1}}
\usepackage{tikz}
\usepackage{multirow}
\usepackage{svg}
\usepackage{eucal}
\usepackage{lipsum}
\usepackage{algorithm}
\usepackage{algorithmic}
\usepackage{amsmath}

\usepackage{booktabs}

\usepackage{multirow}

\usepackage{algorithm}
\usepackage{algorithmic}
\usepackage{amsmath}
\usepackage{booktabs}

\usepackage{multirow}

\usepackage{newfloat}
\usepackage{listings}

\usepackage{soul}
\usepackage{xcolor}
\definecolor{marine}{RGB}{157,195,230}
\definecolor{orange}{RGB}{255,192,0}
\definecolor{lightpink}{HTML}{EA33F7}
\definecolor{lightgreen}{HTML}{D2F6D7}
\definecolor{lightblue}{HTML}{D7FAED}
\definecolor{mypink}{HTML}{FFD4D8}
\definecolor{paleblue}{HTML}{D5EAF6}
\definecolor{palered}{HTML}{F7C4C0}

\usepackage{amsfonts}

\usepackage{amsmath}

\begin{document}
\title{Structural Attention: Rethinking Transformer for Unpaired Medical Image Synthesis}
%
%
\author{Vu Minh Hieu Phan\inst{1} \and
Yutong Xie
\and
Bowen Zhang
\and
Yuankai Qi
\and
Zhibin Liao
\and
Antonios Perperidis
\and
Son Lam Phung
\and
Johan W. Verjans
\and
Minh-Son To}



\maketitle              
\begin{abstract}
Unpaired medical image synthesis aims to provide complementary information for an accurate clinical diagnostics, and address challenges in obtaining aligned multi-modal medical scans. Transformer-based models excel in imaging translation tasks thanks to their ability to capture long-range dependencies. Although effective in supervised training settings, their performance falters in unpaired image synthesis, particularly in synthesizing structural details. 
This paper empirically demonstrates that, lacking strong inductive biases,  Transformer can converge to non-optimal solutions in the absence of paired data. To address this, we introduce UNet Structured Transformer (UNest) — a novel architecture incorporating structural inductive biases for unpaired medical image synthesis. We leverage the foundational Segment-Anything Model to precisely extract the foreground structure and perform structural attention within the main anatomy. 
This guides the model to learn key anatomical regions, thus improving structural synthesis under the lack of supervision in unpaired training. 
Evaluated on two public datasets, spanning three modalities, \textit{i.e.}, MR, CT, and PET, UNest improves recent methods by up to 19.30\% across six medical image synthesis tasks. 
\end{abstract}
\section{Introduction}
Multi-modal medical imaging, including modalities such as computed tomography (CT), magnetic resonance imaging (MRI) and positron emission tomography (PET), can be a valuable tool with multiple applications in clinical practice, aiding in accurate disease diagnosis~\cite{cui2023deep,doherty2013midbrain}, lesion detection~\cite{dai2021transmed} and treatment planning~\cite{learning2022radiomics}. While these modalities provide complementary insights, obtaining multiple scans can be time-consuming, costly, and, most importantly, potentially harmful to patients through additional radiation exposure~\cite{richardson2015risk} (i.e., CT and PET). Medical image synthesis has emerged as a potential solution to enable comprehensive patient evaluation without requiring multiple scans. 

Most synthesis methods employ a supervised Pix2Pix~\cite{emami2021sa,dalmaz2022resvit,zhang2022map}, 
showing effectiveness given paired data between two domains. 
Yet, acquiring substantial paired data is challenging due to variations in resolution and patient positioning.
This highlights the advantage of unpaired image synthesis. CycleGAN~\cite{zhu2017unpaired} is a pioneering work in unpaired image translation, which imposes domain cycle consistency using two generators. Subsequent approaches further incorporate structural consistency loss, either by aligning pixel-wise correlations~\cite{yang2020unsupervised,yang2018unpaired,matsuo2022unsupervised} or preserving shape fidelity~\cite{tang2021attentiongan,ge2019unpaired} between original and synthesized images. Previous methods adopt convolution operators with the local inductive biases, guiding the model to extract local features.  
This limits their ability to capture long-range spatial contexts. 

\begin{figure}[!h]
    \centering
    \includegraphics[width=0.7\linewidth]{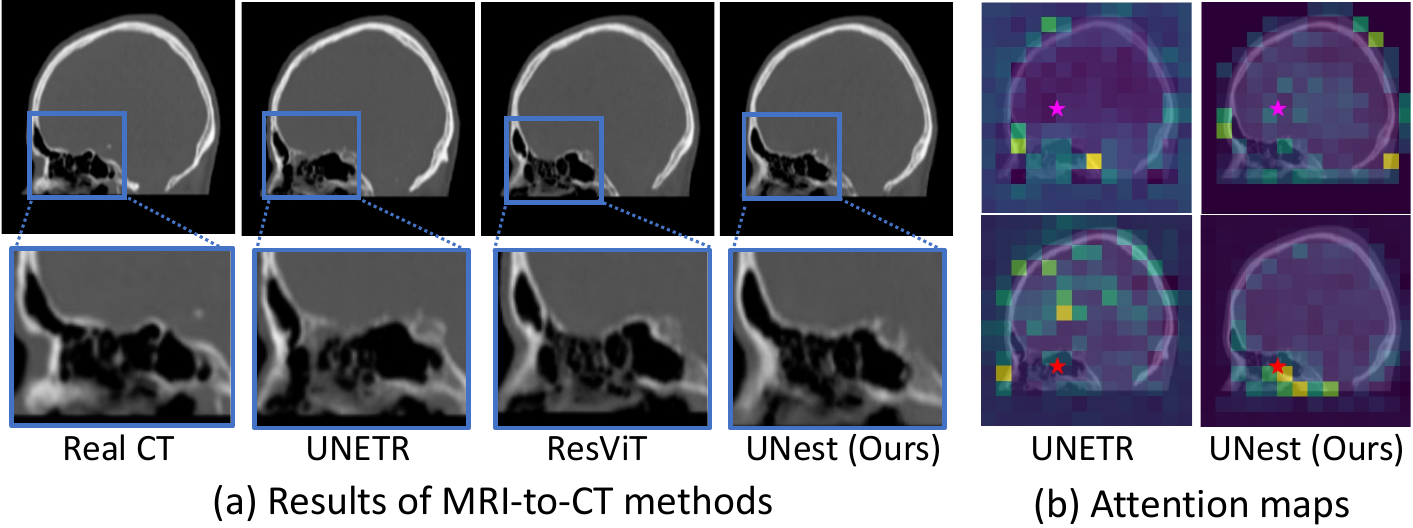}
    \caption{(a) Synthetic MR-to-CT results of different ViT methods: ResViT~\cite{dalmaz2022resvit}, UNETR~\cite{hatamizadeh2022unetr}, and our proposed UNest, where our UNest most accurately preserves the structural carvity. (b) Attention maps for two patches in a smooth brain region (highlighted by a \textcolor{lightpink}{purple} star) a structural nasal cavity (indicated by a \textcolor{red}{red} star). Transformer methods tend to focus on less relevant background features. 
    }
    \label{fig:teaser}
\end{figure}


Recently, Vision Transformer (ViT)~\cite{dosovitskiy2020image} discards the inductive biases to learn features globally. It has achieved state-of-the-art results for many supervised image-to-image translation tasks such as medical semantic segmentation~\cite{hatamizadeh2022unetr,tang2022self}, and  image super-resolution~\cite{chen2023activating,yu2023osrt}. 
ResViT~\cite{dalmaz2022resvit} has advanced the use of hybrid Convolution-Transformer models for paired image translation.  While several unpaired methods~\cite{ristea2023cytran,torbunov2023uvcgan} develop hybrid models, their architectures primarily rely on convolutional operators. 

Unfortunately, ViT models~\cite{hatamizadeh2022unetr,dalmaz2022resvit} struggle when applied to unpaired medical image synthesis. Fig.~\ref{fig:teaser}(a) shows examples that current ViT methods fail to synthesize intricate anatomical structures in the nasal cavities. 
Without convolution-like inductive biases, recent studies~\cite{lu2022bridging,raghu2021vision} show that ViT is less sample-efficient and fail to attend to discriminative features in low data regime.  
In this paper, we empirically discover that the ViT's performance is non-optimal in unpaired image translation scenarios,
where the lack of supervision signals exaggerates the problem. 
Fig.~\ref{fig:teaser}(b) shows that UNETR and ResViT attend to less relevant background regions when learning foreground structures in the brain and nasal cavity (respectively denoted by the \textcolor{lightpink}{pink} and \textcolor{red}{red} stars), suggesting non-optimal solution under unpaired training. 
Solutions such as Swin UNETR~\cite{tang2022self} use local attention on specific windows; yet, confining the context within a restricted window can exclude crucial anatomical regions.

This paper introduces a strong inductive bias into the Transformer and proposes a simple-yet-effective architecture, called UNEt Structural Transformer (UNest) for effective unpaired medical image synthesis. At the core of UNest, our Structural Transformer (ST) block segregates foreground and background tokens, and injects structural and local inductive biases respectively on each region.
When synthesizing foreground, our structural attention aggregates contexts within the anatomy to encode relationships between anatomical regions. 
Fig.~\ref{fig:teaser}(b) shows that UNest adaptively focuses on local areas when synthesizing structural cavities (\textcolor{lightpink}{pink}) and more globally 
when generating smooth brain (\textcolor{red}{red}). 

Our contributions can be summarized as follows. \textbf{1)} We show empirically that injecting a structural inductive bias enables Transformer to focus on discriminative areas, thus enhancing the synthesis of anatomical structures in unpaired image synthesis. \textbf{2)} We introduce a simple-yet-effective architecture, coined UNet Structural Transformer (UNest), applying a dual attention strategy: structural attention for the foreground and local attention for the background. \textbf{3)} Evaluated across six translation tasks covering three modalities: MR, CT, and PET, our method significantly improves the accuracy across various anatomical structures, from head to whole-body images.
\section{Method}
\begin{figure}[!h]
    \centering
    \includegraphics[width=0.9\linewidth]{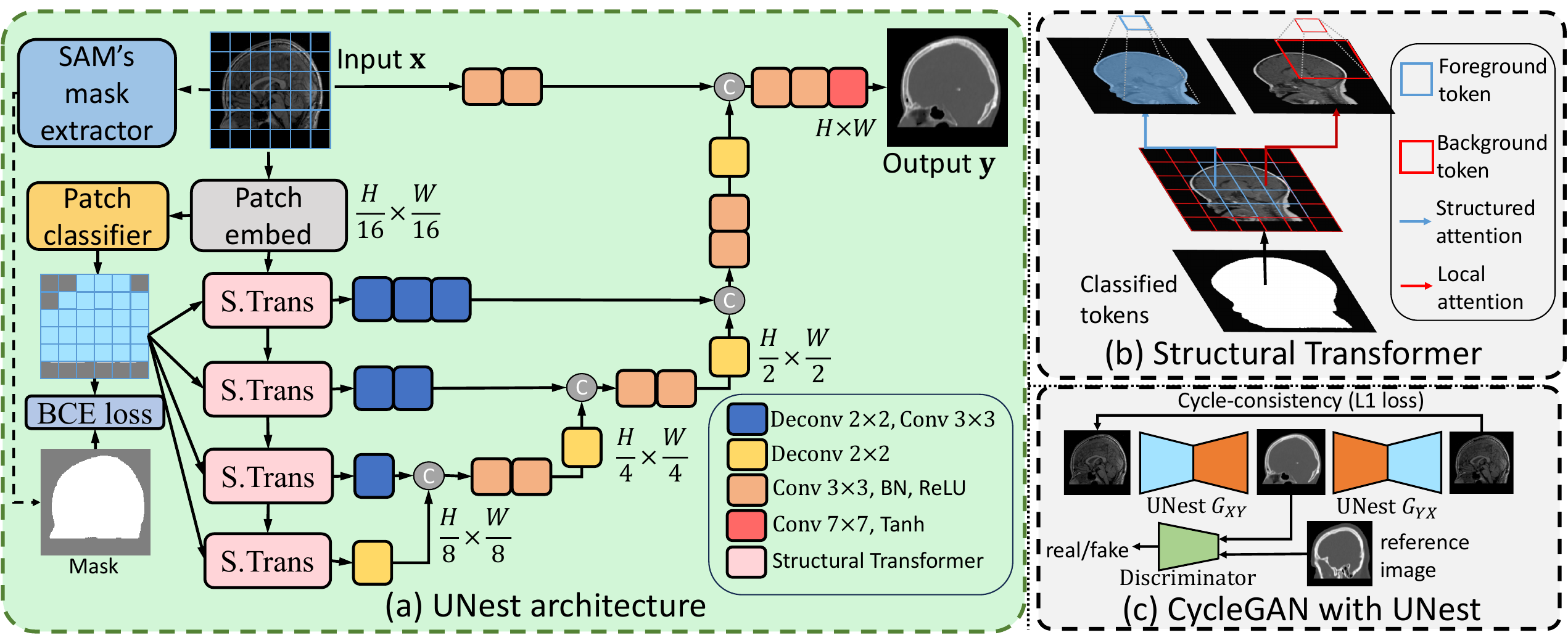}
    \caption{(a) UNest architecture 
    uses Structural Transformer blocks (\sethlcolor{mypink}\hl{pink} block), as shown in (b), to perform dual-attention strategy on foreground and background tokens separately. The decoder upsamples features using deconvolutional layers and skip-connections from early encoder layers. (c) CycleGAN with UNest generators. }
    \label{fig:method}
\end{figure}

\noindent \textbf{CycleGAN's overview.} 
Let $\{\mathbf{x}_n\}_{i=1}^n$ and $\{\mathbf{y}_m\}_{m=1}^M$ 
denote unpaired training samples from domains $X$ and $Y$.
Based on CycleGAN~\cite{zhu2017unpaired}, our framework consists of two generators $G_{XY}: X \mapsto Y$ and $G_{YX}: Y \mapsto X$, learning forward and backward mappings between the two domains, as shown in Fig.~\ref{fig:method}(c). 
Here, $G_{XY}$ and $G_{YX}$ are trained to fool the respective discriminator $D_Y$ and $D_X$ via adversarial loss
\begin{equation}
    \mathcal{L}_{\text{adv,Y}}(G_{\text{X}}, D_{\text{Y}}) = \mathbb{E}_{\mathbf{y} \sim p(\mathbf{y})} \big[ \log D_{\text{Y}}(\mathbf{y}) \big] + \mathbb{E}_{\mathbf{x} \sim p(\mathbf{x})} \big[ \log (1- D_{\text{Y}}(G_{\text{X}}(\mathbf{x}))).
\end{equation}
Adversarial loss $\mathcal{L}_{\text{adv,X}}$ for training  generator $G_Y$ is defined similarly. For unpaired training, CycleGAN imposes the cycle consistency loss 
\begin{equation}
    \mathcal{L}_{\text{cycle}} = \mathbb{E}_{\mathbf{x} \sim p(\mathbf{x})} \|\mathbf{x} - G_{Y}(G_{X}(\mathbf{x}))\| + \mathbb{E}_{\mathbf{y} \sim p(\mathbf{y})} \|\mathbf{y} - G_{X}(G_{Y}(\mathbf{y}))\|.
\end{equation}

\noindent \textbf{Analysis of Transformer models on unpaired image synthesis.} Given a feature map $F \in \mathbb{R}^{H \times W \times K}$, the attention mechanism in Transformer~\cite{vaswani2017attention} first projects a feature map $F$ into query $Q$, key $K$, and value $V$. Given a query token $\mathbf{q} \in \mathbb{R}^K$, it aggregates the contextual information from a set of $T$ surrounding key-value tokens $K_S \in \mathbb{R}^{T \times K}$, and $V_S \in \mathbb{R}^{T \times K}$. 
The global attention methods (e.g., UNETR~\cite{hatamizadeh2022unetr}, ResViT~\cite{dalmaz2022resvit}) consider aggregating the entire image with a global scope $S$: 
\begin{equation}
    S=\{(i,j) | i \in [1,H], j \in [1, W]\}.
\end{equation}
Learning on a global scope, Transformer is considered lacking local inductive bias. In contrast, the local attention methods (e.g., Swin UNETR) incorporate a local inductive bias, inspired by CNN. They aggregate tokens from a local $m \times m$ window 
surrounding the query token:
\begin{equation}
    S=\{(i,j) | i \in [i_q - \frac{m}{2}, i_q + \frac{m}{2}], j \in [j_q - \frac{m}{2}, j_q + \frac{m}{2}]\},
\end{equation}
where $(i_q, j_q)$ is the index of the query token $q$ in the image. The self-attention module $\text{SA}(\cdot, \cdot)$ learns a new query feature by aggregating the key $K_S$ and value vectors $V_S$ within the pre-defined scope $S$ as $\tilde{F} = \text{SA}(F, S) = \text{Softmax}(QK^T_S) V_S$.

\begin{figure}[!h]
    \centering
    \includegraphics[width=0.9\linewidth]{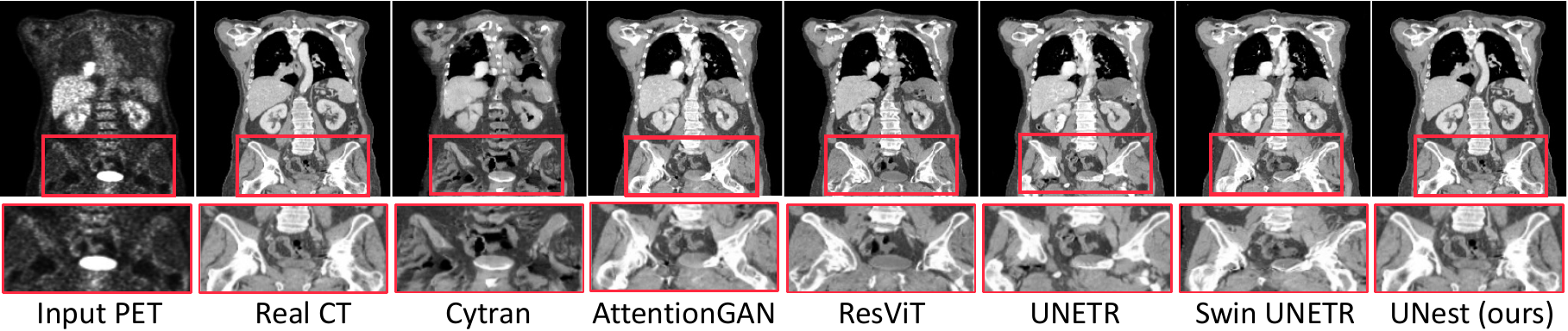}
    \caption{Visual results for PET-to-CT on AutoPET dataset~\cite{gatidis2022whole}.}
    \label{fig:pet-ct}
\end{figure}

Despite achieving high accuracy on supervised learning tasks~\cite{dalmaz2022resvit,zhang2022ptnet3d,zhang2022segvit,hatamizadeh2022unetr}, 
current Transformer models falter under unpaired image synthesis. Fig.~\ref{fig:pet-ct} shows that global attention adopted in ResViT~\cite{dalmaz2022resvit} and UNETR~\cite{hatamizadeh2022unetr} deforms the hip structures, while the local attention in Swin UNETR~\cite{tang2022self} generates artifacts. 
This indicates that lacking strong inductive biases makes current Transformer models converge to a non-optimal solution under unpaired training.

\noindent \textbf{Our UNet Structural Transformer.} Given the lack of supervision in unpaired image translation, we propose to inject the structural inductive bias, enforcing the attention on discriminative anatomical structures. We introduce a 
UNest architecture, for unpaired medical image synthesis.  UNest consists of Structural Transformer (ST) blocks
and a convolutional decoder with skip connections, as shown in Fig.~\ref{fig:method}(a).

As shown in Fig.~\ref{fig:method}(a), Transformer-based encoder $E(\cdot)$ first splits the input image $\mathbf{x}$ of shape $H \times W$ into a sequence of $HW/p^2$ 
non-overlapping patches with patch size $p\times p$. We flatten this sequence and denote it as $X_p \in \mathbb{R}^{(H' \cdot W') \times P}$, where $H'=\frac{H}{p}$, $W'=\frac{W}{p}$, and $P=p^2$ is the total size of each
patch token. 
Each patch token $\mathbf{x}^i_p \in \mathbb{R}^{P}$ is projected  into a $K$ dimensional embedding space via linear layer $\theta \in \mathbb{R}^{P \times K}$: 
\begin{equation}
    \mathbf{f}_0 = [\mathbf{x}^1_p \theta, \mathbf{x}^2_p \theta,\ldots, \mathbf{x}^{H'W'}_p \theta].
\end{equation}

Different from previous global~\cite{hatamizadeh2022unetr,chen2021transunet} and local attention~\cite{tang2022self}, our structural attention aggregates contexts within the main anatomy. In other words, the query scope $S$ consists of tokens in the foreground anatomy. To realize this, we
use a
lightweight patch classifier to classify each token as foreground or background. The classifier uses a $1 \times 1$ convolution, followed by a Sigmoid function. The result is a binary mask $M \in \mathbb{R}^{H' \times W'}$, where each element $M_{h,w}$ represents the probability of the token at position $(h,w)$ being foreground. 


To train the patch classifier, the ground-truth binary mask $\hat{M}$ is extracted 
by leveraging foundational Segment-Anything Model (SAM)~\cite{kirillov2023segment}. As SAM is a class-agnostic segmentation, we use it to extract all masks in the image and choose the one covering the largest foreground area, indicative of the primary anatomy. 
We compute the ground-truth masks offline. During training, they are loaded to compute the binary cross-entropy (BCE) loss between ground-truth and predicted masks to optimize the patch classifier:

\begin{equation}
    \mathcal{L}_{\text{mask}} = -\frac{1}{H' \times W'}\displaystyle\sum_{i=1}^{H'}\displaystyle\sum_{j=1}^{W'}\left[\hat{M}_{ij}\log(M_{ij}) + (1 - \hat{M}_{ij})\log(1 - M_{ij})\right].
\end{equation}

\noindent The decoder receives the contextual features from ST blocks and generates the synthetic image $\tilde{\mathbf{y}}$ with skip connections. Each decoder layer gradually upsamples encoded features using a sequence of $2 \times 2$ deconvolution and a $3 \times 3$ convolution. 

The final generator loss combines the CycleGAN's loss~\cite{zhu2017unpaired}, including cycle consistency and adversarial losses, and the mask loss, weighted by $\lambda$:
\begin{equation}
    \mathcal{L} = \mathcal{L}_{\text{adv}} +  \mathcal{L}_{\text{cycle}} + \lambda \mathcal{L}_{\text{mask}}.
\end{equation}


\noindent \textbf{Details of our Structural Transformer block.} To guide the Transformer under unpaired training, our Structural Transform applies dual attention strategy. For the foreground, we perform structural attention to learn relationships between anatomical regions. For the background, the local attention is performed, enabling effective information exchange between foreground and background features.
The ST block uses the binary mask $M$ to classify foreground tokens given a threshold $\sigma$. 
For a query token $q_{\text{fg}}$ in the foreground, our model aggregates key-value tokens within a scope $S_{\text{fg}}$ within the foreground anatomy:
\begin{equation}
    S_{\text{fg}} = \{(i, j) \mid M_{i,j} > \sigma \}.
\end{equation}
A default value $\sigma=0.5$ is used to classify foreground tokens. The structural attention consists of a layer normalization (\text{LN}) followed by a multi-head self-attention (\text{MSA}) layer~\cite{vaswani2017attention} and a multi-layer perceptron (\text{MLP}) with skip connections to facilitate the preservation of low-level features. The foreground token output $\mathbf{f}^{\text{fg}}_{t+1}$ from the structural attention in the $t$-th ST block is formulated as:
\begin{align}
  \mathbf{f'}^{\text{fg}}_{t+1} &= \text{SA}(\text{LN}(\mathbf{f}^{\text{fg}}_t), S_{\text{fg}}) + \mathbf{f}^{\text{fg}}_t, \text{ and} \\
  \mathbf{f}^{\text{fg}}_{t+1} &= \text{MLP}(\text{LN}(\mathbf{f'}^{\text{fg}}_{t+1})) + \mathbf{f'}^{\text{fg}}_{t+1}.
\end{align}

For a query token $q_{\text{bg}}$ in the background, we aggregate tokens around a local window $m \times m$:
\begin{equation}
    S_{\text{bg}} = \{(i,j) | i \in [i_q - \frac{m}{2}, i_q + \frac{m}{2}], j \in [j_q - \frac{m}{2}, j_q + \frac{m}{2}]\}.
\end{equation}
Similarly, the background token output $\mathbf{f}^{\text{bg}}_{t+1}$ from the local attention is formulated as:
\begin{align}
  \mathbf{f'}^{\text{bg}}_{t+1} &= \text{SA}(\text{LN}(\mathbf{f}^{\text{bg}}_t), S_b) + \mathbf{f}^{\text{bg}}_t, \text{ and}  \\
  \mathbf{f}^{\text{bg}}_{t+1} &= \text{MLP}(\text{LN}(\mathbf{f'}^{\text{bg}}_t)) + \mathbf{f'}^{\text{bg}}_t.
\end{align}

\noindent Lastly, the dual-attention ST block consolidates the output $\mathbf{f}_{t+1}$ based on the indices $(h,w)$ designated for the foreground $S_{\text{fg}}$, and the background scope $S_{\text{bg}}$.

\section{Experimental Results}
\noindent \textbf{Datasets and implementation details.}
For MR-to-CT and MR-to-PET translation tasks, experiments are conducted on the MRXFDG~\cite{merida2021cermep} dataset, a public neuroimaging set containing cerebral scans from 37 adult subjects. 
Sagittal slices are extracted and resized to $224 \times 224$ pixels. 
For the PET-to-CT translation task, models are evaluated on the \textit{AutoPET} dataset~\cite{gatidis2022whole}, comprising 310 whole-body PET and CT scans acquired from a Siemens Biograph mCT scanner. Coronal views are extracted and resized to $256 \times 256$ pixels. Both datasets are partitioned into training, validation, and test sets with the proportion of 8:1:1. 

 All models are trained using the Adam optimizer for 100 epochs, with a learning rate of 1e-4 which linearly decays to zero over the last 50 epochs. We train with a batch size of 16 on two NVIDIA RTX 3090 GPUs. 

\noindent \textbf{Evaluation metrics.} Employing standard performance metrics from prior studies~\cite{yang2020unsupervised,liu2021ct}, we compute mean absolute error (MAE),  peak signal-to-noise ratio (PSNR), and structural similarity (SSIM) between the real and synthesized images. The reported results are averaged over five runs. A paired student's $t$-test is conducted between UNest and the compared methods to test the significance of performance difference ($p=0.05$).

\begin{table}[!h]
    \centering
    \caption{Comparison of different methods on head scan, MRXFDG dataset~\cite{merida2021cermep} for four translation tasks. The improvement of UNest over all compared methods is statistically significant at $p=0.05$.}
    \label{tab:sota-mr-pet}
    \resizebox{0.99\textwidth}{!}{%
    \begin{tabular}{@{}lccc||ccc||ccc||ccc@{}}
    \toprule \multirow{ 2}{*}{\textbf{Methods}} & \multicolumn{3}{c}{\textbf{MR-to-PET}} & \multicolumn{3}{c}{\textbf{PET-to-MR}} & \multicolumn{3}{c}{\textbf{MR-to-CT}} & \multicolumn{3}{c}{\textbf{CT-to-MR}} \\
         & MAE $\downarrow$ & PSNR $\uparrow$ & SSIM $\uparrow$ & MAE $\downarrow$ & PSNR $\uparrow$ & SSIM $\uparrow$ & MAE $\downarrow$ & PSNR $\uparrow$ & SSIM $\uparrow$ & MAE $\downarrow$ & PSNR $\uparrow$ & SSIM $\uparrow$ \\\midrule
         \textit{Convolution-based} \\ \midrule
         CycleGAN~\cite{zhu2017unpaired} & 7.88$_{\pm .1}$ & 33.11$_{\pm .01}$ & 79.53$_{\pm .2}$ & 11.21$_{\pm .1}$ & {31.89}$_{\pm .02}$ & 62.47$_{\pm .3}$ & 7.20$_{\pm .1}$ & 34.67$_{\pm .04}$ & 82.56$_{\pm .5}$ & 10.75$_{\pm .2}$ & 32.15$_{\pm .06}$ & 70.54$_{\pm .7}$ \\
         sc-CycleGAN~\cite{yang2020unsupervised} & 7.54$_{\pm .1}$ & 33.20$_{\pm .02}$ & 80.06$_{\pm .3}$ & 10.64$_{\pm .2}$ & {31.93}$_{\pm .03}$ & 62.87$_{\pm .3}$ & 7.12$_{\pm .2}$ & 34.82$_{\pm .05}$ & 83.06$_{\pm .7}$ & 10.34$_{\pm .4}$ & 32.33$_{\pm .11}$ & 70.77$_{\pm .8}$  \\
AttentionGAN~\cite{tang2021attentiongan} & 7.21$_{\pm .1}$ & 33.54$_{\pm .03}$ & 81.09$_{\pm .2}$ & 10.23$_{\pm .1}$  & 33.02$_{\pm .02}$ & 70.97$_{\pm .1}$ & 6.84$_{\pm .1}$ & 34.97$_{\pm .05}$ & 83.88$_{\pm .6}$ & 9.92$_{\pm .4}$ & 33.12$_{\pm .09}$ & 72.09$_{\pm .9}$ \\
MaskGAN~\cite{phan2023structure} & 6.95$_{\pm .1}$ & 33.98$_{\pm .02}$ & 81.76$_{\pm .2}$ & 9.81$_{\pm .1}$  & 33.10$_{\pm .02}$ & 71.35$_{\pm .2}$ & 6.59$_{\pm .2}$ & 35.10$_{\pm .05}$ & 84.53$_{\pm .7}$ & 9.37$_{\pm .3}$ & 33.39$_{\pm .09}$ & 72.53$_{\pm 1.0}$ \\
          \midrule
         \textit{Hybrid Conv-Trans} \\ \midrule 
        TransUNet~\cite{chen2021transunet} & 13.10$_{\pm .2}$ & 32.68$_{\pm .04}$ & 70.81$_{\pm .6}$ & 13.70$_{\pm .3}$ & 31.78$_{\pm .03}$ & 59.25$_{\pm .5}$  & 10.25$_{\pm .2}$ & 33.47$_{\pm .04}$ & 75.52$_{\pm .8}$ & 12.48$_{\pm .3}$ & 31.22$_{\pm .06}$ & 62.32$_{\pm .9}$ \\
         ResViT~\cite{dalmaz2022resvit} & 10.44$_{\pm .1}$ & 32.99$_{\pm .03}$ & 76.84$_{\pm .5}$ & 14.66$_{\pm .2}$ & 30.62$_{\pm .03}$ & 56.24$_{\pm .4}$ & 7.58$_{\pm .2}$ & 34.72$_{\pm .03}$ & 81.89$_{\pm .7}$ & 11.72$_{\pm .3}$ & 31.85$_{\pm .07}$ & 64.51$_{\pm 1.3}$   \\
        Cytran~\cite{ristea2023cytran} & 8.11$_{\pm .1}$ & 33.23$_{\pm .02}$ & 78.29$_{\pm .5}$ & 11.05$_{\pm .3}$ & 32.22$_{\pm .04}$ & 60.74$_{\pm .5}$ & 6.98$_{\pm .1}$ & 34.85$_{\pm .04}$ & 82.68$_{\pm .7}$ & 9.85$_{\pm .2}$ & 32.45$_{\pm .06}$ & 70.80$_{\pm 1.0}$ \\ %

         \midrule
         \textit{Pure Transformers} \\ \midrule
         PTNet~\cite{zhang2022ptnet3d} & 11.43$_{\pm .3}$ & 32.76$_{\pm .07}$ & 70.53$_{\pm .8}$ & 13.63$_{\pm .4}$ & 32.33$_{\pm .08}$ & 63.21$_{\pm .9}$ & 9.42$_{\pm .4}$ & 33.90$_{\pm .10}$ & 75.88$_{\pm .9}$ & 12.08$_{\pm .7}$ & 31.43$_{\pm .13}$ & 62.89$_{\pm 1.2}$ \\  
         UNETR~\cite{hatamizadeh2022unetr} & 7.35$_{\pm .1}$ & 33.60$_{\pm .04}$ & 81.90$_{\pm .4}$ & 8.98$_{\pm .1}$ & 33.06 $_{\pm .03}$& 72.18$_{\pm .2}$ & 7.95$_{\pm .4}$ & 34.55$_{\pm .11}$ & 81.18$_{\pm .9}$ & 10.45$_{\pm .3}$ & 32.98$_{\pm .05}$ & 68.33$_{\pm 1.0}$  \\
         Swin-UNETR~\cite{tang2022self} & 7.54$_{\pm .2}$ & 33.53$_{\pm .03}$ & 81.72$_{\pm .6}$ &  9.11$_{\pm .2}$ & 33.03$_{\pm .07}$ & 71.49$_{\pm .8}$ & 8.80$_{\pm .7}$ & 34.28$_{\pm .18}$ & 80.53$_{\pm 1.11}$ & 11.15$_{\pm .5}$ & 32.87$_{\pm .05}$ & 67.93$_{\pm 1.3}$\\
         \midrule
         UNest (Ours) & \textbf{6.32}$_{\pm .2}$ & \textbf{34.24}$_{\pm .04}$ & \textbf{82.97}$_{\pm .4}$ & \textbf{8.10}$_{\pm .3}$ & \textbf{33.91}$_{\pm .05}$ & \textbf{74.13}$_{\pm .3}$ & \textbf{6.18}$_{\pm .3}$ & \textbf{35.76}$_{\pm .09}$ & \textbf{85.48}$_{\pm .5}$ & \textbf{8.72}$_{\pm .4}$ & \textbf{34.17}$_{\pm .07}$ & \textbf{72.97}$_{\pm .9}$ \\\hline
    \end{tabular}%
    }
\end{table}

\begin{figure}[!h]
    \centering
    \includegraphics[width=0.85\linewidth]{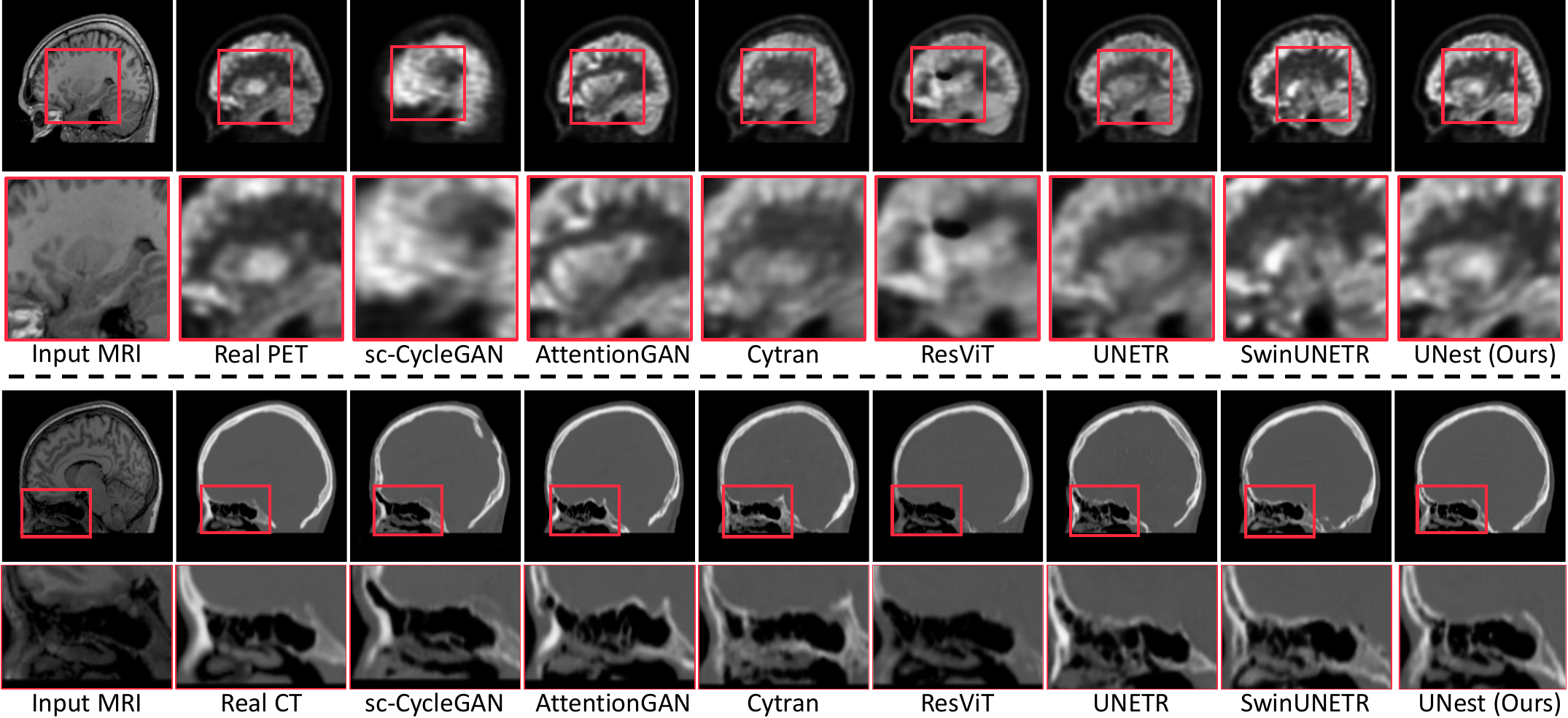}
    \caption{Visual results of different methods for MRI-to-PET and MRI-to-CT translation on MRXFDG dataset.}
    \label{fig:mri-pet}
\end{figure}

\noindent \textbf{Comparisons with state-of-the-art.} Table~\ref{tab:sota-mr-pet} benchmarks four synthesis tasks 
on the MRXFDG dataset. 
Without \textit{inductive bias}, UNETR is inferior to convolution AttentionGAN on a small-sized dataset, which is common in the medical domain. 
Injecting strong learning biases, the proposed UNest outperforms Swin-UNETR by 19.30\% and 3.95\% respectively in terms of MAE and PSNR. 
Fig.~\ref{fig:mri-pet} presents visual results on MR-to-PET and MR-to-CT. Without inductive bias, UNETR tends to produce blurrier details, while Swin-UNETR distorts the details in the subcortical brain structures. 
In contrast, our structural attention method, accurately synthesizes the 
structural regions in PET. Table~\ref{tab:sota-pet-ct} presents the benchmark of different synthesis methods on AutoPET dataset~\cite{gatidis2022whole} for PET-to-CT and CT-to-PET tasks. Incorporating a structural inductive bias, our UNest significantly improves upon UNETR and Swin-UNETR by 14.21\% and 11.73\% respectively in terms of MAE.

\begin{table}
    \centering
        \caption{Quantitative comparison of different methods on the AutoPET dataset for translating between PET and CT. The improvement of UNest over all compared methods is statistically significant at $p=0.05$. FG-S + BG-S: an ablated version of UNest with structural attention applying on both foreground and background regions.}
    \label{tab:sota-pet-ct}
    \resizebox{0.85\textwidth}{!}{%
    \begin{tabular}{@{}lccc||ccc@{}}
    \toprule \multirow{ 2}{*}{\textbf{Methods}} & \multicolumn{3}{c}{\textbf{PET-to-CT}} & \multicolumn{3}{c}{\textbf{CT-to-PET}} \\
         & MAE $\downarrow$ & PSNR $\uparrow$ & SSIM $\uparrow$ & MAE $\downarrow$ & PSNR $\uparrow$ & SSIM $\uparrow$  \\\midrule
         \textit{Convolution-based} \\ \midrule
         CycleGAN~\cite{zhu2017unpaired} & 14.34$\pm.2$ & 32.30$\pm.04$ & 66.85$\pm .6$ & 12.80$\pm.1$ & 31.95$\pm.02$ & 69.53$\pm.5$ \\
         sc-CycleGAN~\cite{yang2020unsupervised} & {14.59}$\pm .2$ & {32.42}$\pm .06$ & {67.43}$\pm .8$ & {12.77}$\pm .1$ & {32.05}$\pm .04$ & 70.53$\pm .8$ \\
AttentionGAN~\cite{tang2021attentiongan} & {13.62}$\pm .2$ &{32.97}$\pm .06$ & {67.12}$\pm .6$ & {11.87}$\pm .1$  &{32.61}$\pm .03$ & 72.54$\pm .6$ \\
         QS-Attention~\cite{hu2022qs} & {13.21}$\pm .4$  & {33.70}$\pm .08$ & 71.49$\pm1.1$ & 9.75$\pm.2$ & 33.21$\pm.07$ & 77.42$\pm 1.0$ \\
          \midrule
         \textit{Hybrid Convolution-Transformer} \\ \midrule 
        TransUNet~\cite{chen2021transunet} & 16.65$\pm .5$ & 32.68$\pm.10$ & 62.68$\pm1.1$ & 13.52$\pm.4$ &  31.94$\pm.09$ & 67.88$\pm 1.1$  \\
        Cytran~\cite{ristea2023cytran} & 14.77$\pm .3$ & 32.24$\pm.08$ & 64.37$\pm.9$ & 9.05$\pm.2$ & 33.56$\pm.06$ & 74.65$\pm.8$ \\ 
         ResViT~\cite{dalmaz2022resvit} & 13.55$\pm.3$ & 33.28$\pm.07$ & 70.53$\pm.7$ & 9.90$\pm.1$ & 32.13$\pm.06$ & 68.29$\pm.7$  \\
         \midrule
         \textit{Pure Transformers} \\ \midrule
         PTNet~\cite{zhang2022ptnet3d} & 14.11$\pm.6$ & 32.62$\pm.10$ & 66.53$\pm1.3$ & 9.24$\pm.5$ & 32.77$\pm.09$ & 70.53$\pm1.1$ \\
         UNETR~\cite{hatamizadeh2022unetr} & 11.08$\pm.3$ & 33.70$\pm.08$ & 76.06$\pm.8$ & 7.77$\pm.2$ & 33.58$\pm.05$ & 80.30$\pm.8$ \\
         Swin-UNETR~\cite{tang2022self} & 10.93$\pm.4$ & 33.78$\pm.08$ & 76.42$\pm.9$ & 7.42$\pm.3$ & 33.78$\pm.07$ & 80.53$\pm.8$ \\
         \midrule
         UNest (Ours) & \textbf{9.57}$\pm.3$  & \textbf{34.05}$\pm.07$ & \textbf{78.47}$\pm.8$ & \textbf{6.55}$\pm.2$ & \textbf{34.27}$\pm.05$ & \textbf{81.55}$\pm.7$ \\
         UNest w/ FG-S + BG-S & 9.73 $\pm .4$ & 33.85 $\pm .07$ & 77.85 $\pm .8$ & 6.87 $\pm .3$ & 34.11 $\pm .05$ & 81.21 $\pm .8$ \\\hline
    \end{tabular}%
    }
\end{table}



\noindent \textbf{Ablation study.} Table~\ref{tab:sota-pet-ct} shows that using dual attention with structural attention FG-S + BG-S improves MAE of UNETR and Swin UNETR by 12.18\% and 10.98\%, respectively, on PET-CT. Separating attentions on BG and FG injects a structural inductive bias, useful for unpaired training. 
\textit{Notably}, using our hybrid attention with FG-S and BG-L achieves the best results, which is further supported by visual results in Fig.~\ref{fig:att-abl}-Left. Fig.~\ref{fig:att-abl}-Right shows that global attention attends to less relevant BG tokens, while our structural attention adaptively attends to anatomical features. It incorporates a long-range context for smooth brain regions (top) and a localized focus for a structural sinus (bottom). 

\definecolor{myyellow}{HTML}{BF9000}
\definecolor{myblue}{HTML}{2E75B6}
\begin{figure}[!h]
    \centering
    \includegraphics[width=0.55\linewidth]{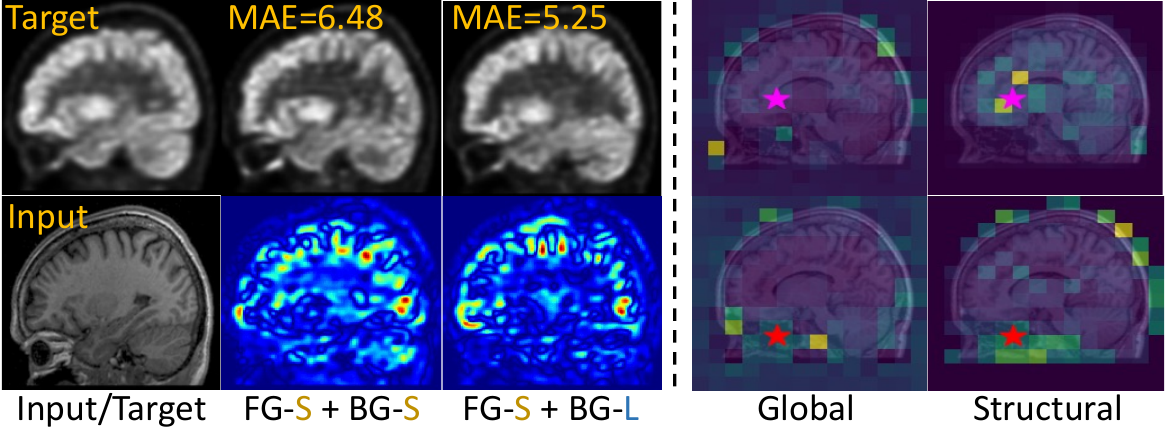}
    \caption{\textit{Left}: Error maps of synPET produced by UNest with FG-\textcolor{myyellow}{S} + BG-\textcolor{myyellow}{S} attention, and our hybrid \textcolor{myyellow}{S}-\textcolor{myblue}{L} attention. \textit{Right}: Attention maps of global and structural attention.}
    \label{fig:att-abl}
\end{figure}

\section{Conclusion}
In this paper, we introduce the UNet Structural Transformer (UNest) for unpaired medical image synthesis. Traditional Transformer methods often underperform in unpaired image synthesis, due to the absence of appropriate learning priors. Injecting structural priors as in our UNest significantly reduces the synthesis error of previous Transformer models by 19.3\% on the small-sized MRXFDG. The structural inductive bias also boosts synthesis performance of current models by 14.2\% on the large-scale AutoPET dataset. 

\bibliographystyle{splncs04}
\bibliography{paper456}

\end{document}


\title{Supplementary Material}
%
%

\authorrunning{Anonymous}
%
%
%
\maketitle              

\begin{figure}[!h]
    \centering
\includegraphics[width=\linewidth]{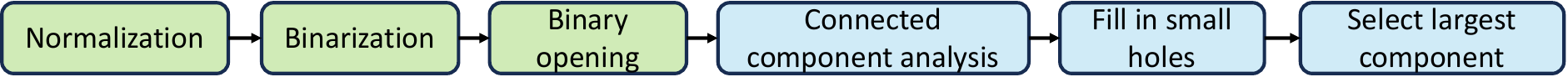}
    \caption{Traditional image processing for mask extraction consists of two stages: image pre-processing (\textcolor{green}{green} blocks) and connected component analysis (\textcolor{blue}{blue} blocks).}
    \label{fig:extract}
\end{figure}

\begin{figure*}[!h]
    \centering
    \includegraphics[width=\linewidth]{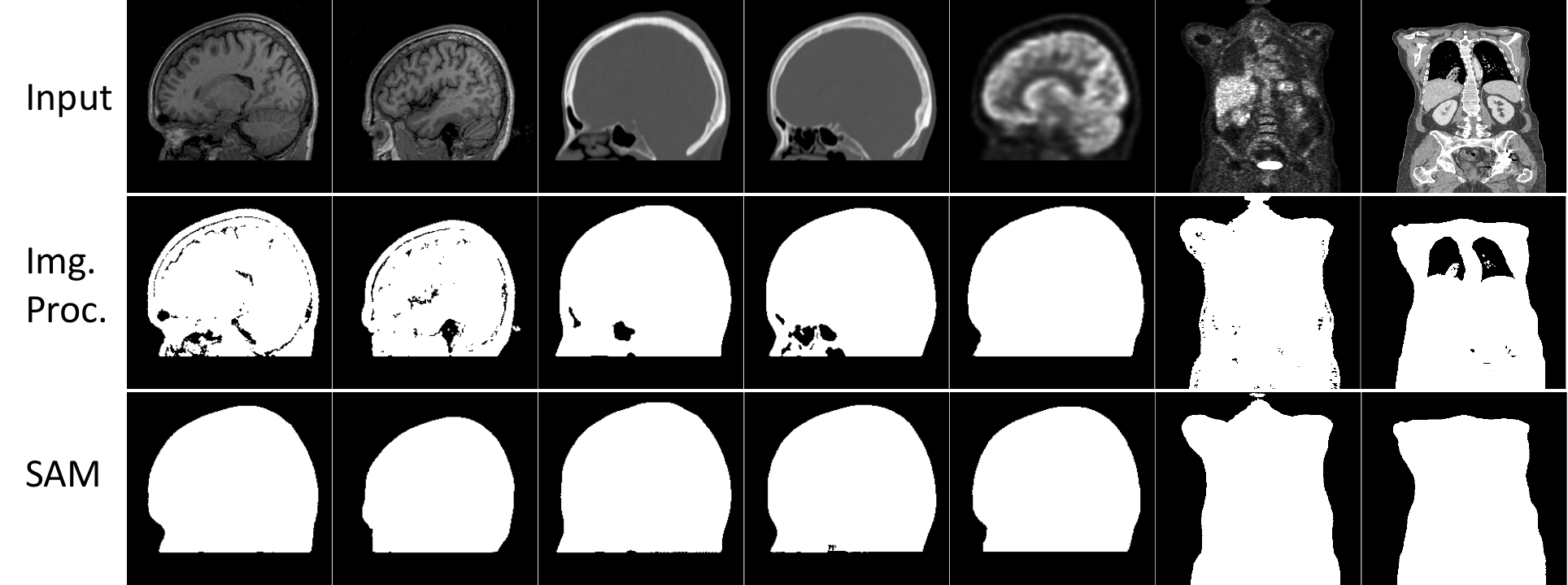}
    \caption{Extracted masks using standard image processing algorithms (Row 2) and using deep-learning-based Segment-Anything-Model (Row 3).}
    \label{fig:mask}
\end{figure*}


\begin{table}[!h]
    \centering
    \caption{Accuracies on three tasks using different attention mechanisms for foreground (FG) and background (BG).  Reversing the attention strategies with structural attention on BG reduces the performance significantly.}
    \label{tab:abl-switch-att}
    \begin{tabular}{@{}cc||cccccc@{}}
    \toprule 
    \multirow{2}{*}{FG} & \multirow{2}{*}{BG} & \multicolumn{2}{c}{PET-CT} & \multicolumn{2}{c}{MR-CT} & \multicolumn{2}{c}{MR-PET} \\\cmidrule(lr){3-4}\cmidrule(lr){5-6}\cmidrule(lr){7-8}
    &  & MAE$\downarrow$ & SSIM$\uparrow$ & MAE$\downarrow$ & SSIM$\uparrow$ & MAE$\downarrow$ & SSIM$\uparrow$  \\\midrule
    \multicolumn{2}{c||}{Global} & 10.91 & 75.88 & 7.84 & 81.23 & 7.33 & 81.96   \\
    \multicolumn{2}{c||}{Local} & 10.82 & 75.97 & 8.12 & 81.06 & 7.64  & 81.53  \\\midrule
    Local & Struct & 10.51 & 75.91  & 7.75 & 81.48 & 7.21 & 82.15  \\
    Struct & Local  & \textbf{9.57} & \textbf{78.47} & \textbf{6.18} & \textbf{85.48} & \textbf{6.32} & \textbf{82.97} \\\bottomrule
    \end{tabular}
\end{table}

\begin{table}[!h]
    \caption{Hyper-parameter analysis. Evaluating the effect of mask loss weight ($\lambda_{\text{mask}}$), local attention window size (win. size), and classification threshold ($\sigma$) on PET-CT performance. The model performance is insensitive when classification threshold, $\sigma > 0.5$. Using low $\lambda_{\text{mask}}=0.1$ can lead to lower performance, showing the effects of learning structural information via the mask loss. A high window size with a value of 8 leads to optimal performance. Using a large window size for local attention on background allows a more effective information exchange between foreground and background features.}
    \label{tab:hyperparam}
    \centering
    \begin{tabular}{@{}ccc||ccc@{}}
        \toprule
        $\lambda_{\text{mask}}$ & win. size & $\sigma$ & MAE & PSNR & SSIM \\
        \midrule
        0.5 &  2 & 0.5 & 9.84 & 33.89 & 77.70 \\
        0.5 & 4 & 0.5 & 9.71 &  33.96 & 78.12 \\\midrule  
        0.5 & 8 & 0.05 & 10.02 &  33.78 & 76.12 \\
        0.5 & 8 & 0.25 & 9.72 &  33.98 & 78.20 \\
        0.5 & 8 & 0.6 & 9.65 &  34.01 & 78.31 \\
        0.5 & 8 & 0.8 & 9.66 &  34.03 & 78.38 \\\midrule
        0.1 &  8 & 0.5 & 9.92 & 33.86 & 77.48 \\
        0.5 & 8 & 0.5 & \textbf{9.57} & \textbf{34.05} & \textbf{78.47} \\
        1.0 & 8 & 0.5 & 9.75 & 33.93 & 78.06 \\
        \bottomrule
    \end{tabular}
\end{table}

\begin{table}[!h]
    \centering
    \caption{Ablation study on learning masks via a patch classifier versus using ground-truth (GT). GT masks are extracted via image processing or SAM. Plus, stop-gradient (SG) from a mask loss to a patch classifier is ablated.  Inference time on $224 \times 224$ images is reported. Performance of using GT masks from SAM is similar to using the patch classifier, while increasing computational overhead (from 0.014s to 1.924s), showing effectiveness of our design for \textit{distilling structural knowledge} from a powerful SAM to a lightweight patch classifier. Stopping gradients to patch classifier reduces performance, showing the benefits of training it with mask loss.}
    \label{tab:gt_use}
    \resizebox{0.99\textwidth}{!}{%
    \begin{tabular}{cc|c||c|ccc|ccc|ccc}
        \multirow{2}{*}{Patch cls.} & \multirow{2}{*}{SG} & \multirow{2}{*}{GT mask gen.} & \multirow{2}{*}{Overhead (s)} & \multicolumn{3}{c|}{PET-CT} & \multicolumn{3}{c|}{MR-PET} & \multicolumn{3}{c}{MR-CT} \\\cmidrule(lr){5-7} \cmidrule(lr){8-10} \cmidrule(lr){11-13}
            &    & &  & MAE & PSNR & SSIM & MAE & PSNR & SSIM & MAE & PSNR & SSIM \\\midrule
        \cmark & \cmark & \xmark & \multirow{2}{*}{0.014}  &9.57 & 34.05 & 78.47 & 6.32 & 34.24 & 82.97 & 6.18 & 35.76& 85.48 \\
        \cmark & \xmark & \xmark &   & 9.82 & 33.82 & 77.34 & 6.62 & 34.11 & 81.88 & 6.41 & 35.23 & 84.72 \\\midrule
        \multicolumn{2}{c|}{\xmark} & Img proc. & 0.016 (\tri 0.002) & 10.51 & 33.82 & 77.02 & 6.87 & 34.03 & 81.73 & 7.25 & 34.85 & 83.41 \\
        \multicolumn{2}{c|}{\xmark}& SAM & 1.924 (\tri 1.91) & 9.71 & 33.97 & 78.05 & 6.54 & 34.15 & 82.26 & 6.30 & 35.55 & 84.93 \\\midrule
    \end{tabular}%
    }
\end{table}


